\documentclass[conference]{IEEEtran}
\ifCLASSINFOpdf
\else
\fi

\usepackage{amsmath, amsthm, amssymb}
\usepackage{bm}
\usepackage{bbm}
\usepackage{float}
\usepackage{graphicx}
\usepackage{subfig}
\usepackage{ctable}
\usepackage[]{algorithm2e}

\begin{document}
\title{Binary Classifier Calibration using an Ensemble of Near Isotonic Regression Models}

\maketitle

\begin{abstract}
Learning accurate probabilistic models from data is crucial in many practical tasks in data mining. In this paper we present a new non-parametric calibration method called \textit{ensemble of near isotonic regression} (ENIR). The method can be considered as an extension of BBQ \cite{pakdaman2015obtaining}, a recently proposed calibration method, as well as the commonly used calibration method based on isotonic regression. ENIR is designed to address the key limitation of isotonic regression which is the monotonicity assumption of the predictions. Similar to BBQ, the method post-processes the output of a binary classifier to obtain calibrated probabilities.  Thus it can be combined with many existing classification models. We demonstrate the performance of ENIR on synthetic and real datasets for the commonly used binary classification models. Experimental results show that the method outperforms several common binary classifier calibration methods. 
In particular on the real data, ENIR commonly performs statistically significantly better than the other methods, and never worse. It is able to improve the calibration power of classifiers, while retaining their discrimination power. The method is also computationally tractable for large scale datasets, as it is $O(N \log N)$ time, where $N$ is the number of samples. 
\end{abstract}
\section{Introduction}

This paper focuses on developing a new non-parametric calibration method for post-processing the output of commonly used binary classification models to generate accurate probabilities. Obtaining accurate probabilities is crucial in many real world decision making and data mining problems. Decision theory provides a rationale basis for intelligent agents to make decisions \cite{russell2010artificial}. Decision theory combines utilities and probabilities in determining the actions that maximize expected utility. In general, the probabilities need to be well calibrated in order to achieve this goal of finding the best actions.

\begin{figure}
\centering
         \includegraphics[width=\linewidth]{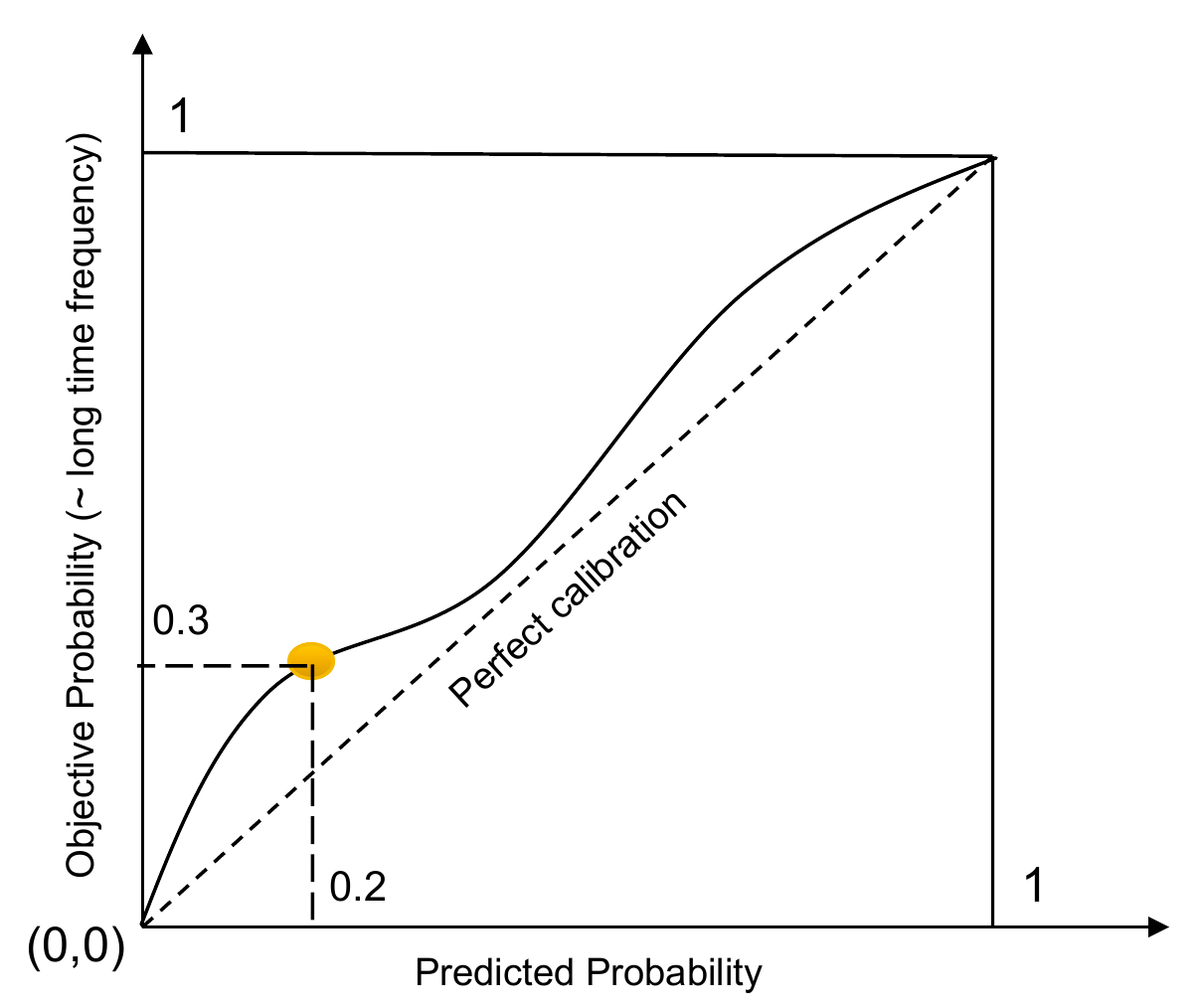}
        %  [scale= 0.64]{./Figures/reliability-curve.png}
         \caption{The solid line shows a calibration (reliability) curve for predicting
        $Z=1$. The dotted line is the ideal calibration curve.}
       \label{IMG:Fig1}
\end{figure}

Informally, we say that a classification model is well calibrated if events predicted to occur with probability \textit{p} do occur about \textit{p} fraction of the time, for all \textit{p}. This concept applies to binary as well as multi-class classification problems. Figure \ref{IMG:Fig1} illustrates the binary calibration problem using a reliability curve \cite{degroot1983comparison,niculescu2005predicting}. The curve shows the probability predicted by the classification model versus the actual fraction of positive outcomes for a hypothetical binary classification problem, where \textit{Z }is the binary event being predicted. The curve shows that when the model predicts $Z=1$ to have probability $0.2$, the outcome $Z=1$ occurs in about $0.3$ fraction of the time. The curve shows that the model is fairly well calibrated, but it tends to underestimate the actual probabilities. In general, the straight dashed line connecting $(0, 0)$ to $(1, 1)$ represents a perfectly calibrated model. The closer a calibration curve is to this line, the better calibrated is the associated prediction model.  Deviations from perfect calibration are very common in practice and may vary widely depending on the binary classification model that is used \cite{pakdaman2015obtaining}.

Producing well-calibrated probabilistic predictions is critical in many areas of science (e.g., determining which experiments to perform), medicine (e.g., deciding which therapy to give a patient), business (e.g., making investment decisions), and many others. 
In data mining problems, obtaining well-calibrated classification models is crucial not only for decision-making, but also for combining output of different classification models \cite{bella2013effect}. It is also useful when we aim to use the output of a classifier not only to discriminate the instances but also to rank them \cite{zhang2004naive, jiang2005learning, hashemi2010application}. Research on learning well calibrated models has not been explored in the data mining literature as extensively as, for example, learning models that have high discrimination (e.g., high accuracy).

There are two main approaches to obtaining well-calibrated classification models. The first approach is to build a classification model that is intrinsically well-calibrated \textit{ab initio}. This approach will restrict the designer of the data mining model by requiring major changes in the objective function (e.g, using a different type of loss function) and could potentially increase the complexity and computational cost of the associated optimization program to learn the model. The other approach is to rely on the existing discriminative data mining models and then calibrate their output using post-processing methods. This approach has the advantage that it is general, flexible, and it frees the designer of a data mining algorithm from modifying the learning procedure and the associated optimization method \cite{pakdaman2015obtaining}. However, this approach has the potential to decrease discrimination while increasing calibration, if care is not taken. 

The method we describe in this paper is shown empirically to improve calibration of different types of classifiers (e.g., LR, SVM, and NB) while maintaining their discrimination performance well. Existing post-processing binary classifier calibration methods include Platt scaling \cite{platt1999probabilistic}, histogram binning \cite{zadrozny2001obtaining}, isotonic regression \cite{zadrozny2002transforming}, and a recently proposed method BBQ which is a Bayesian extension of histogram binning \cite{pakdaman2015obtaining}. In all these methods, the post-processing step can be seen as a function that maps the outputs of a prediction model to probabilities that are intended to be well-calibrated. Figure \ref{IMG:Fig1} shows an example of such a mapping.

In general, there are two main applications of post-processing calibration methods. First, they can be 
used to convert the outputs of discriminative classification methods with no apparent probabilistic 
interpretation to posterior class probabilities \cite{platt1999probabilistic}. An example is an SVM model that learns a discriminative model that does not have a direct probabilistic interpretation. In this paper, we show this use of calibration to map SVM outputs to well-calibrated probabilities. Second, calibration methods can be applied to improve the 
calibration of predictions of a probabilistic model that is miscalibrated. For example, a na\"{i}ve Bayes (NB) 
model is a probabilistic model, but its class posteriors are often miscalibrated due to unrealistic independence 
assumptions \cite{niculescu2005predicting}. The method we describe is shown empirically to improve the calibration 
of NB models without reducing their discrimination. The method can also work well on calibrating models that are 
less egregiously miscalibrated than are NB models.

\section{Related work}

Existing post-processing binary classifier calibration models can be divided into parametric and non-parametric methods. Platt's method is an example of the former; it uses a sigmoid transformation to map the output of a classifier into a calibrated probability \cite{platt1999probabilistic}. The two parameters of the sigmoid function are learned in a maximum-likelihood framework using a model-trust minimization algorithm \cite{gill1981practical}. The method was originally developed to transform the output of an SVM model into calibrated probabilities. It has also been used to calibrate other type of classifiers \cite{niculescu2005predicting}. The method runs in $O(1)$ at test time, and thus, it is fast. Its key disadvantage is the restrictive shape of sigmoid function that rarely fits the true distribution of the predictions \cite{jiang2012calibrating}.

A popular non-parametric calibration method is the equal frequency histogram binning model which is also known as quantile binning \cite{zadrozny2001obtaining}. In quantile binning, predictions are partitioned into $B$ equal frequency bins. For each new prediction $y$ that falls into a specific bin, the associated frequency of observed positive instances will be used as the calibrated estimate for $P(z=1 | y)$, where $z$ is the true label of an instance that is either $0$ or $1$. Histogram binning can be implemented in a way that allows it to be applied to large scale data mining problems. Its limitations include (1) bins inherently “pigeonhole” calibrated probabilities into only $B$ possibilities, (2) bin boundaries remain fixed over all predictions, and (3) there is uncertainty in the optimal number of the bins to use \cite{zadrozny2002transforming}.

The most commonly used non-parametric classifier calibration method in machine learning and data mining applications is the \textit{isotonic regression based calibration} (IsoRegC) model \cite{zadrozny2002transforming}. To build a mapping from the uncalibrated output of a classifier to the calibrated probability, IsoRegC assumes the mapping is an isotonic (monotonic) mapping following the ranking imposed by the base classifier. The commonly used algorithm for isotonic regression is the \textit{Pool Adjacent Violators Algorithm} (PAVA), which is linear in the number of training data \cite{barlow1972statistical}. An IsoRegC model based on PAVA can be viewed as a histogram binning model \cite{zadrozny2002transforming} where the position of the boundaries are selected by fitting the best monotone approximation to the train data according to the ordering imposed by the classifier. There is also a variation of the isotonic-regression-based calibration method for predicting accurate probabilities with a ranking loss \cite{menon2012predicting}. In addition, an extension to IsoRegC combines the outputs generated by multiple binary classifiers to obtain calibrated probabilities \cite{zhong2013accurate}. While IsoRegC can perform well on some real datasets, the monotonicity assumption it makes can fail in real data mining applications. This can  specifically occur when we encounter large scale data mining problems in which we have to make simplifying assumptions to build the classification models. Thus, there is a need to relax the assumption, which is the focus of the current paper.

Adaptive calibration of predictions (ACP) is another extension to histogram binning \cite{jiang2012calibrating}. ACP requires the derivation of a $95\%$ statistical confidence interval around each individual prediction to build the bins. It then sets the calibrated estimate to the observed frequency of the instances with positive class among all the predictions that fall within the bin. To date, ACP has been developed and evaluated using only logistic regression as the base classifier \cite{jiang2012calibrating}.

Recently, a new non-parametric calibration model called BBQ was proposed which is a refinement of the histogram-binning calibration method \cite{pakdaman2015obtaining}. BBQ addresses the main drawbacks of the histogram binning model by considering multiple different equal frequency histogram binning models and their combination using a Bayesian scoring function \cite{heckerman1995learning}. However, BBQ has two disadvantages. First, as a post processing method, BBQ does not utilize a key prior knowledge in calibrating the output of a classifier. This prior knowledge is that, in most real data mining problems, the classifier should perform well in terms of discrimination measures (e.g., AUC measure), otherwise we are not interested in using it. Second, BBQ still selects the position and boundary of the bins by considering only equal frequency histogram binning models. A Bayesian non-parametric method called ABB addresses the later problem by considering Bayesian averaging over all possible binning models induced by the training instances \cite{pakdaman20015binary}. The main drawback of ABB is that it is computationally intractable for most real world applications, as it requires $O(N^2)$ computations for learning the model as well as $O(N^2)$ computations for computing the calibrated estimate for each of the test instances\footnote{Note that the running time for the test instance can be reduced to $O(1)$ in any post-processing calibration model by using a simple caching technique that reduces calibration precision in order to decrease calibration time \cite{pakdaman20015binary}}.

This paper presents a new binary classifier calibration method called \textit{ensemble of near isotonic regression} (ENIR) that can post process the output generated by a wide variety of classification models. The essential idea in ENIR is to use the prior knowledge that the scores that are going to be calibrated are in fact generated by a well performing classifier in terms of discrimination. IsoRegC also uses such prior knowledge; however, it is biased by constraining the calibrated scores to obey the ranking imposed by the classifier. In the limit, this is equivalent to presuming the classifier has AUC equal to $1$, which rarely happens in real world data mining applications. In contrast, BBQ does not make any assumptions about the correctness of classifier rankings. ENIR provides a balanced approach that spans between IsoRegC and BBQ. In particular, ENIR assumes that the mapping from uncalibrated scores to calibrated probabilities is a near isotonic (monotonic) mapping; It allows violations of the ordering imposed by the classifier and then penalizes them through the use of a regularization term. ENIR utilizes the path algorithm \textit{modified pool adjacent violators algorithm} (mPAVA) that can find the solution path to near isotonic regression problem in $O(N \log N)$, where $N$ is the number of training instances \cite{tibshirani2011nearly}. Finally, it uses the BIC scoring measure to combine the predictions made by these models to yield more robust calibrated predictions.

The remainder of this paper is organized as follows. Section \ref{sec:method} introduces the ENIR method. Section \ref{sec:experiments} describes a set of experiments that we performed to evaluate ENIR and other calibration methods. Finally, Section \ref{sec:conclusion} states conclusions and describes several areas for future work.

\section{Method}
\label{sec:method}
In this section we introduce the \textit{ensemble of near isotonic regression} (ENIR) calibration method. ENIR utilizes the near isotonic regression method \cite{tibshirani2011nearly} that seeks a nearly monotone approximation for a sequence of data $y_1,\ldots,y_n$. The proposed calibration model extends the commonly used isotonic regression based calibration by a (approximate) selective Bayesian averaging of a set of nearly isotonic regression models. The set includes the isotonic regression model as an extreme member. From another viewpoint, ENIR can be considered as an extension to a recently proposed calibration model BBQ \cite{ pakdaman2015obtaining} by relaxing the assumption that probability estimates are independent inside the bins and finding the boundary of the bins automatically through an optimization algorithm. 

Before getting into the details of the method, we define some notation. Let $y_i$ and $z_i$ define respectively an uncalibrated classifier prediction and the true class of the $i$'th instance. In this paper, we focus on calibrating a binary classifier's output\footnote{For classifiers that output scores that are not in the unit interval (e.g. SVM), we use a simple sigmoid transformation $f(x) = \frac{1}{1 + \exp(-x)}$ to transform the scores into the unit interval.}, and thus, $z_i \in \{0,1\}$ and $y_i \in [0,1]$. Let $\mathcal{D}$ define the set of all training instances $(y_i , z _i)$. Without loss of generality, we can assume that the instances are sorted based on the classifier scores $y_i$, so we have $y_1 \le y_2 \le \ldots \le y_N$, where $N$ is the total number of samples in the training data. 

The standard isotonic regression based calibration model finds the calibrated probability estimates by solving the following optimization problem:

\begin{equation}
\begin{aligned}
& \hat{\bm{p}}_{iso} = \underset{\bm{p} \in R^N}{\text{argmin}}
& & \frac{1}{2} \sum_{i=1}^{N} (p_i-z_i)^2 \\
& \text{s.t.} & &  p_1 \le \ldots \le p_N \\
& & &  0\le p_i \le 1 \; \forall i \in \{1, \ldots, N\}, \\
\end{aligned}
\label{eq:iso}
\end{equation}

where $\hat{\bm{p}}_{iso}$ is the vector of calibrated probability estimates. The rationale behind the model is to assume that the base classifier ranks the instances correctly. To find the calibrated probability estimates, it seeks the best fit of the data that are consistent with the classifier's ranking. A unique solution to the above convex optimization program exists and can be obtained by an inductive iterative algorithm called \textit{pool adjacent violator algorithm} (PAVA) that runs in $O(N)$. Note, however, that isotonic regression calibration still needs $O(N \log N)$ computations, due to the fact that instances are required to be sorted based on the classifier scores $y_i$. PAVA iteratively groups the consecutive instances that violate the ranking constraint and uses their average over $z$ (frequency of positive instances) as the calibrated estimate for all the instances within the group. We define the set of these consecutive instances that are located in the same group and attain the same predicted calibrated estimate, as a bin. Therefore, an isotonic regression-based calibration can be viewed as a histogram binning method \cite{ zadrozny2002transforming} where the position of boundaries are selected by fitting the best monotone approximation to the training data according to the ranking imposed by the classifier. 

One can show that the second constraint in the optimization given by Equation \ref{eq:iso} is redundant, and it is possible to rewrite the equation in the following equivalent form:

\begin{equation}
\begin{aligned}
& \hat{\bm{p}}_{iso} = \underset{\bm{p} \in R^N}{\text{argmin}}
& &\frac{1}{2} \sum_{i=1}^{N} (p_i-z_i)^2 + \lambda \sum_{i=1}^{N-1} (p_i - p_{i+1}) \nu_i\\
& \text{s.t.} & &  \lambda = + \infty,
\end{aligned}
\end{equation}

where $\nu_i = \mathbbm{1}(p_i > p_{i+1})$ is the indicator function of ranking violation. Relaxing the equality constraint in the above optimization program leads to a new optimization problem called nearly isotonic regression \cite{tibshirani2011nearly}.

\begin{equation}
\begin{aligned}
& \hat{\bm{p}}_{\lambda} = \underset{\bm{p} \in R^N}{\text{argmin}}
& &\frac{1}{2} \sum_{i=1}^{N} (p_i-z_i)^2 + \lambda \sum_{i=1}^{N-1} (p_i - p_{i+1})\nu_i,
\end{aligned}
\label{eq:neariso}
\end{equation}
where $\lambda$ is a positive real number that regulates the tradeoff between the monotonicity of the calibrated estimates with the goodness of fit by penalizing adjacent pairs that violate the ordering imposed by the base classifier. The above optimization problem is convex having a unique solution $\hat{\bm{p}_{\lambda}}$, where the use of the subscript $\lambda$ emphasizes the dependency of the final solution to the value of $\lambda$.

The entire path of solutions for any value of $\lambda$ of the near isotonic regression problem can be found using a similar algorithm to PAVA which is called \textit{modified pool adjacent violator algorithm} (mPAVA) \cite{tibshirani2011nearly}. mPAVA finds the whole solution path in $O(N \log N)$, and needs $O(N)$ memory space. Briefly, the algorithm works as follows: It starts by constructing $N$ bins, each bin containing a single instance of the train data. Next, it finds the solution path by starting from the saturated fit $p_i = z_i$, that corresponds to setting $\lambda = 0$, and then increasing $\lambda$ iteratively. As the $\lambda$ increases the calibrated probability estimates $\hat{\bm{p}}_{\lambda, i}$, for each bin, will change linearly with respect to $\lambda$ until the calibrated probability estimates of two consecutive bins attain equal value. At this stage, mPAVA merges the two bins that have the same calibrated estimate to build a larger bin, and it updates their corresponding estimate to a common value. The process continues until there is no change in the solution for a large enough value of $\lambda$ that corresponds to finding the standard isotonic regression solution. The essential idea of mPAVA is based on a theorem stating that if two adjacent bins are merged on some value of $\lambda$ to construct a larger bin, then the new bin will never split for all larger values of $\lambda$ \cite{tibshirani2011nearly}.

mPAVA yields a collection of nearly isotonic calibration models, with the over fitted calibration model at one end ($\hat{\bm{p}}_{\lambda=0} =\bm{z}$) and the isotonic regression solution at the other ($\hat{\bm{p}}_{\lambda= \lambda_\infty} = \hat{\bm{p}}_{iso}$), where $\lambda_\infty$ is a large positive real number. Each of these models can be considered as a histogram binning model where the position of boundaries and the size of bins are selected according to how well the model trades off the goodness of fit with the preservation of the ranking generated by the classifier, which is governed by the value of $\lambda$, (As $\lambda$ increases the model is more concerned to preserving the original ranking of the classifier, while for the small $\lambda$ it prioritizes the goodness of fit.)

ENIR employs the approach just described to generate a collection of models (one for each value of $\lambda$). It then uses the Bayesian Information Criterion (BIC) to score each of the models \footnote{Note that we exclude the highly overfitted model that corresponds to $\lambda = 0$ from the set of models in ENIR}. Assume mPAVA yields the binning models $M_1, M_2, \ldots, M_T$, where $T$ is the total number of models generated by mPAVA. For any new classifier output $y$, the calibrated prediction in the \textit{ENIR} model is defined using selective Bayesian model averaging \cite{hoeting1999bayesian}:
\begin{equation*}
\label{Eq-BayesianAveraging}
\begin{split}
    P(z=1|y)  =\sum_{i=1}^{T}  \frac{ Score(M_i)}{\sum_{j=1}^{T} Score(M_j)} P(z=1|y,M_i) ,
\end{split}
\end{equation*}
where $P(z=1|y,M_i)$ is the probability estimate obtained using the binning model $M_i$ for the uncalibrated classifier output $y$. Also, $Score(M_i)$ is defined using the BIC scoring function \cite{ schwarz1978estimating}.
% $T$ is the total number of binning models considered and
Next, for the sake of the completeness, we briefly describe the mPAVA algorithm; more detailed information about the algorithm and the derivations can be found in \cite{tibshirani2011nearly}.

% \footnote{Note that, as it is recommended in \cite{tibshirani2011nearly}, we use the expected degree of freedom of the nearly isotonic regression models, which is equivalent to the number of bins, as the number of parameters in the BIC scoring function.}
\subsection{The modified PAV algorithm}

Suppose at a value of $\lambda$ we have $N_\lambda$ bins, $B_1, B_2, \ldots,B_{N_\lambda}$. We can represent the unconstrained optimization program given by Equation \ref{eq:neariso} as the following loss function that we seek to minimize :

\begin{equation}
\begin{aligned}
& \mathcal{L}_{B,\lambda}(\bm{z}, \bm{p}) =
&&\frac{1}{2} \sum_{i=1}^{N_\lambda} \sum_{j \in B_i} (p_{B_i} - z_i)^2 + \\
&&&\lambda \sum_{i=1}^{N_\lambda - 1} (p_{B_i} - p_{B_{i+1}}) \nu_i,
\end{aligned}
\end{equation}
where $p_{B_i}$ defines the common estimated value for all the instances located at the bin $B_i$. The loss function $\mathcal{L}_{B,\lambda}$ is always differentiable with respect to $p_{B_i}$ unless two calibrated probabilities are just being joined (which only happens if $p_{B_i}=p_{B_{i+1}}$ for some $i$). Assuming that $\hat{p}_{B_i}(\lambda)$ is optimal, the partial derivative of $\mathcal{L}_{B,\lambda}$ has to be $0$ at $\hat{p}_{B_i}(\lambda)$, which implies:

\begin{equation}
|B_i|\hat{p}_{B_i}(\lambda) - \sum_{j \in B_i} z_j  + \lambda (\nu_i - \nu_{i-1})=0 \text{  for } i = 1,\ldots, N_\lambda
\label{eq:loss_derivative}
\end{equation}
% & \frac{\partial{\mathcal{L}_{B,\lambda}}}{\partial{\bm{p_{B_i}}}} =

Rewriting the above equation, the optimum predicted value for each bin can be calculated as:

\begin{equation}
\hat{p}_{B_i}(\lambda)= \frac{\sum_{j \in B_i} z_j - \lambda \nu_i + \lambda \nu_{i-1}}{|B_i|}  \text{  for } i = 1,\ldots, N_\lambda
\label{eq:opt_prob}
\end{equation}

While PAVA uses the frequency of instances in each bin as the calibrated estimate, Equation \ref{eq:opt_prob} shows that mPAVA uses a shrunken version of the frequencies by considering the estimates that are not following the ranking imposed by the base classifier. In Equation \ref{eq:loss_derivative}, taking derivatives with respect to $\lambda$ yields:

\begin{equation}
\frac{\partial{\hat{p}_{B_i}}}{{\partial{\lambda}}} = \frac{\nu_{i-1}-\nu_i}{|B_i|}, \text{   for } i = 1,\ldots, N_\lambda, 
\label{eq:slope}
\end{equation}

where we set $\nu_0 = \nu_N = 0$ for notational convenience. As we noted above, it has been proven that the optimal values of the instances located in the same bin are tied together and the only way that they can change is to merge two bins as they can never split apart as the $\lambda$ increases \cite{tibshirani2011nearly}. Therefore, as we make changes in $\lambda$, the bins $B_i$, and hence the values $\nu_i$ remain constant. This implies the term $\frac{\partial{\hat{p}_{B_i}}}{{\partial{\lambda}}}$ is a constant in Equation \ref{eq:slope}. Consequently, the solution path remains piecewise linear as $\lambda$ increases, and the breakpoints happen when two bins merge together. Now, using the piecewise linearity of the solution path and assuming that the two bins $B_i$ and $B_{i+1}$ are the first two bins to merge by increasing $\lambda$, the value of $\lambda_{i,i+1}$ at which the two bins $B_i$ and $B_{i+1}$ will merge is calculated as:

\begin{equation}
\lambda_{i,i+1} = \frac{\hat{p}_{B_i}(\lambda) - \hat{p}_{B_{i+1}}(\lambda)}
{a_{i+1}-a_{i}} + \lambda \text{  for  } i= 1, \ldots N_\lambda -1 ,
\label{eq:lambda}
\end{equation}

where $a_{i} = \frac{\partial{\hat{p}_{B_i}}}{\partial{\lambda}}$ is the slope of the changes of $\hat{p}_{B_i}$ with respect to $\lambda$ according to Equation \ref{eq:slope}. Using the above identity, the $\lambda$ at which the next breakpoint occurs is obtained using the following equation:

\begin{equation}
\begin{aligned}
&\lambda^* = \underset{i}{\text{ min  }} \lambda_{i,i+1}\\
&\mathbbm{I}^* = \{i| \lambda_{i,i+1} = \lambda^* \},
\label{eq:lambda-star}
\end{aligned}
\end{equation}

where $\mathbbm{I}^*$ indicates the set of the indexes of the bins that will be merged by their consecutive bins changing the $\lambda$\footnote{Note that there could be more than one bin achieving the minimum in Equation \ref{eq:lambda-star}, so they should be all merged with the bins that are located next to them.}. If $\lambda^* < \lambda$  then the algorithm will terminate since it has obtained the standard isotonic regression solution, and by increasing $\lambda$ none of the existing bins will ever merge. Having the solutions of the near isotonic regression problem in Equation \ref{eq:neariso} at the breakpoints, and using the piecewise linearity property of the solution path, it is possible to recover the solution for any value of $\lambda$ through interpolation. However, the current implementation of ENIR only uses the near isotonic regression based calibration models that corresponds to the value of $\lambda$ at the breakpoints. The sketch of the algorithm is shown as Algorithm [\ref{Alg1}].

\begin{algorithm}
 \SetKwInOut{Input}{input}
 \SetKwInOut{Output}{output}
 \SetKwInOut{DownTo}{downto}
 \SetKwInOut{Invariant}{Invariant}
 \Input{$D=\{(y_1,z_1),\ldots,(y_N,z_N)\}$}
 \Output{A set of binning models $M_{1},\ldots, M_{T}$}
 \Invariant{Pairs are sorted based on $y_i$}
 $\lambda \leftarrow 0$\;
 $t \leftarrow 1$\;
 $N_\lambda = N$\; 
% \text{  and  } N_{\lambda}>1$
 \For{$i \leftarrow 1$ \KwTo N}{
    $B_i = \{i\}$  \;
    $p_i = z_i$  \; 
 }
 $\lambda^* \leftarrow \lambda + 1$\;
 \While{$\lambda^* > \lambda$ }{
    Update the slopes $a_i$ using Equation \ref{eq:slope}\;
    Update merging values $\lambda_{i,i+1}$ using Equation \ref{eq:lambda}\;
    Compute $\lambda^*$ and $\mathbbm{I}^*$ using Equation \ref{eq:lambda-star}\;
    \If{$\lambda^* < \lambda$}{
        terminate \;
    } 
    
    \For{$i \leftarrow 1$ \KwTo $N_\lambda$}{
        //update corresponding probability estimate as: \\
        $\hat{p}_{B_i}(\lambda^*) = \hat{p}_{B_i}(\lambda) + a_i \times (\lambda^* - \lambda)$\;
    }
    Merge appropriate bins as indicated in the set $\mathbbm{I}^*$ \;
    Update number of bins $N_\lambda$\;
    Store the corresponding calibration model in $M_t$;\
    $\lambda  \leftarrow \lambda^*$\;
    $t \leftarrow t + 1$ ;\
 }
\caption{The \textit{modified pool adjacent violator algorithm} (mPAVA) that yields a set of near-isotonic-regression-based calibration models $M_1, \ldots, M_T$}
\label{Alg1}
\end{algorithm}

\section{Empirical Results}
\label{sec:experiments}

This section describes the set of experiments that we performed to evaluate the performance of $ENIR$ in comparison to Isotonic Regression based Calibration (IsoRegC) \cite{zadrozny2002transforming}, and a state of the art binary classifier calibration method called BBQ \cite{pakdaman2015obtaining}. We use IsoRegC because it is one of the most commonly used calibration models showing promising performance on real world applications \cite{niculescu2005predicting,zadrozny2002transforming}. Moreover $ENIR$ is an extension of IsoRegC, and we are interested in evaluating whether it performs better than IsoRegC. We also include BBQ as a state of the art binary classifier calibration model, which is a Bayesian extension of the simple histogram binning model \cite{pakdaman2015obtaining}. We did not include Platt's method since it is a simple and restricted parametric model and there are prior works showing that IsoRegC and BBQ perform superior to Platt's method \cite{niculescu2005predicting,zadrozny2002transforming,pakdaman2015obtaining}. We also did not include the ACP method since it requires not only probabilistic predictions, but also a statistical confidence interval ($CI$) around each of those predictions, which makes it tailored to specific classifiers, such as LR \cite{jiang2012calibrating}; this is counter to our goal of developing post-processing methods that can be used with any existing classification models. Finally, we did not include ABB in our experiments mainly because it is not computationally tractable for real datasets that have more than couple of thousands instances. Moreover, even for small size datasets, we noticed that ABB performs quite similarly to BBQ. To evaluate the performance of the methods, we ran experiments on both simulated and on real data.

\subsection{Evaluation Measures}
In order to evaluate the performance of the calibration models, we use $5$ different evaluation measures. We use Accuracy (Acc) and \textit{area under ROC curve} (AUC) to evaluate how well the methods discriminate the positive and negative instances in the feature space. We also utilize the three measures of \textit{root mean square error} (RMSE), \textit{maximum calibration error} (MCE), and \textit{expected calibration error} (ECE) to measure the calibration \cite{pakdaman2015obtaining, pakdaman20015binary}. MCE and ECE are two simple statistics of the reliability curve (Figure \ref{IMG:Fig1} shows a hypothetical example of such curve) computed by partitioning the output space of the binary classifier, which is the interval $[0,1]$, into $K$ fixed number of bins ($K = 10$ in our experiments). The estimated probability for each instance will be located in one of the bins. For each bin we can define the associated calibration error as the absolute difference between the expected value of predictions and the actual observed frequency of positive instances. The $MCE$ calculates the maximum calibration error among the bins, and $ECE$ calculates expected calibration error over the bins, using empirical estimates as follows: 

\begin{equation}
\begin{aligned}
& MCE =  \max_{k=1}^K \left( \left| o_k-e_k \right| \right) \notag \\
& ECE =  \sum_{k=1}^{K} P(k) \cdot \left|o_k-e_k \right|, \notag\\
\end{aligned}
\label{eq:mce-ece}
\end{equation}

where $P(k)$ is the empirical probability or the fraction of all instances that fall into bin $k$, $e_k$ is the mean of the estimated probabilities for the instances in bin $k$, and $o_k$ is the observed fraction of positive instances in bin $k$. The lower the values of $MCE$ and $ECE$, the better is the calibration of a model.

\subsection{Simulated Data} 
\begin{figure}[t]
\centering
\includegraphics[width=\columnwidth]{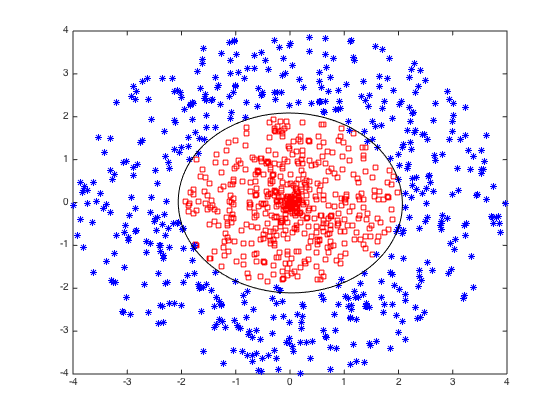}
\caption{Scatter plot of the simulated data. The black oval indicates the decision boundary found using SVM with a quadratic kernel.}
\label{fig:SimulationData}
\end{figure}

For the simulated data experiments, we used a binary classification dataset that was used in previous works \cite{pakdaman2015obtaining,pakdaman20015binary}. The scatter plot of the simulated dataset is shown in Figure \ref{fig:SimulationData}. Pakdaman M. et al. developed this classification problem, which illustrates how IsoRegC can suffer from a violation of the isotonicity assumption. We repeated the experiments in \cite{pakdaman2015obtaining,pakdaman20015binary} to compare the performance of IsoRegC with our new method that assumes approximate isotonicity. In our experiments, the data is divided into $1000$ instances for training and calibrating the prediction model, and $1000$ instances for testing the models. We report the average of $10-$fold cross validation results for the simulated dataset.

To conduct the experiments with the simulated data, we used two extreme classifiers: \textit{support vector machines} (SVM) with linear and quadratic kernels. The choice of SVM with a linear kernel allows us to see how ENIR perform when the classification model makes an over simplifying (linear) assumption. Also, to achieve good discrimination on the circular configuration data in Figure \ref{fig:SimulationData}, SVM with a quadratic kernel is a reasonable choice (as is also evidenced qualitatively in Figure \ref{fig:SimulationData} and quantitatively in Table \ref{tab:circular_svm_quadratic}). So, the experiment using quadratic kernel SVM allows us to see how well ENIR performs when we use models that should discriminate well.

As seen in Table \ref{tab:circular}, ENIR generally outperforms IsoRegC on the simulation dataset, especially when the linear SVM method is used as the base learner. This is due to the monotonicity assumption of IsoReC which presumes the best calibrated estimates will match the ordering imposed by the base classifier. When we use SVM with a linear kernel, this assumption is violated due to the non-linarity of the data. Consequently, IsoRegC only provides limited improvement of the calibration and discrimination performance of the base classifier. ENIR performs very well in this case since it is using the ranking information of the base classifier, but it is not anchored to it. The violation of the monotonicity assumption can happen in real data as well, especially in large scale data mining problems in which we use simple classification models due to the computational constraints. As shown in Table \ref{tab:circular_svm_quadratic}, even when we apply a highly appropriate SVM classifier to classify the instances for which IsoRegC is expected to perform well (and indeed does so), ENIR performs as well or better than IsoRegC.

\begin{table}
\centering
\subfloat[SVM Linear Kernel]{
			\small
            \begin{tabular}{@{}l@{}cccccc@{}}
               \toprule
                 & SVM   & IsoReg & BBQ   & ENIR \\
               \midrule
                AUC   & 0.52  & 0.65  & 0.85  & 0.85 \\
                ACC   & 0.64  & 0.64  & 0.78  & 0.79 \\
                RMSE  & 0.52  & 0.46  & 0.39  & 0.38 \\
                ECE   & 0.28  & 0.35  & 0.05  & 0.05 \\
                MCE   & 0.78  & 0.60  & 0.13  & 0.12 \\
                % LL    & 0.52  & 0.54  & 0.39  & 0.39 \\
               \bottomrule
            \end{tabular}%
           \label{tab:circular_svm_linear}
}\\
\subfloat[SVM Quadratic Kernel]{
			\small
            \begin{tabular}{@{}l@{}cccccc@{}}
             \toprule
               & SVM   & IsoReg & BBQ   & ENIR \\
             \midrule
              AUC   & 1.00  & 1.00  & 1.00  & 1.00 \\
              ACC   & 0.99  & 0.99  & 0.99  & 0.99 \\
              RMSE  & 0.21  & 0.09  & 0.10  & 0.09 \\
              ECE   & 0.14  & 0.01  & 0.01  & 0.00 \\
              MCE   & 0.36  & 0.04  & 0.05  & 0.03 \\
            %   LL    & 0.20  & 0.03  & 0.04  & 0.03 \\
             \bottomrule
          \end{tabular}%
          \label{tab:circular_svm_quadratic}
}
\caption{Experimental Results on a simulated dataset}
\label{tab:circular}
\end{table}

\subsection{Real Data} We used $40$ different baseline real data sets from the UCI and LibSVM repositories\footnote{The datasets used were as follows: spect, adult, breast, pageblocks, pendigits, ad, mamography, satimage, australian, code rna, colon cancer, covtype, letter unbalanced, letter balanced, diabetes, duke, fourclass, german numer, gisette scale, heart, ijcnn$1$, ionosphere scale, liver disorders, mushrooms, sonar scale, splice, svmguide$1$, svmguide$3$, coil$2000$, balance, breast cancer, leu, w$1$a, thyroid sick, scene, uscrime, solar, car$34$, car$4$ , protein homology.} \cite{Bache+Lichman:2013,chang2011libsvm}. Five summary statistics of the size of the datasets and the percentage of the minority class are shown in Table \ref{tab:data_summary}. 

We used three common classifiers, Logistic Regression (LR), Support Vector Machines (SVM), and Na\"{i}ve Bayes (NB) to evaluate the performance of the proposed calibration method. In the experiments we used the average over $10$ random runs of $10$-fold cross validation, and we always used the train data for calibrating the models. To compare the performance of the calibration models, we used the statistical test procedure recommended by Demsar \cite{demvsar2006statistical}. More specifically, we used the Freidman non-parametric hypothesis testing method \cite{friedman1937use} followed by Holm's step-down procedure \cite{holm1979simple} to evaluate the performance of ENIR in comparison with IsoregC and BBQ, across the $40$ baseline datasets.
 
% Table generated by Excel2LaTeX from sheet 'Real Data'
\begin{table}[htbp]
  \centering
  \small
    \begin{tabular}{rccccc}
    \toprule
          & Min   & Q1    & Meadian & Q3    & Max \\
    \midrule
    \multicolumn{1}{l}{Size} & 42    & 683   & 1861  & 8973  & 581012 \\
    \multicolumn{1}{l}{Percent} & 0.009 & 0.076 & 0.340 & 0.443 & 0.500 \\
    \bottomrule
    \end{tabular}%
  \caption{Summary statistics of the size of the real datasets and the percentage of the minority class. Q1 and Q3 defines the first quartile and thirds quartile respectively.}
  \label{tab:data_summary}%
\end{table}%

Tables [\ref{tab:LR},\ref{tab:SVM},\ref{tab:NB}] show the results of the performance of $ENIR$ in comparison with IsoRegC and BBQ. In these tables, we show the average rank of each method across the baseline datasets, where boldface indicates the best performing method. In these tables, the marker $\ast$/$\circledast$ indicates whether ENIR is statistically superior/inferior to the compared method using the Friedman test followed by Holm's step-down procedure at a $0.05$ significance level. For instance, Table \ref{tab:SVM} shows the performance of the calibration models when we use SVM as the base classifier; the results show that ENIR achieves the best performance in terms of RMSE by having an average rank of $1.675$ across the $40$ baseline datasets. The result indicates that in terms of RMSE, ENIR is statistically superior to BBQ; however, it is not performing statistically differently than IsoRegC.

Table \ref{tab:LR} shows the results of comparison when we use LR as the base classifier. As shown, the performance of ENIR is always superior to BBQ and IsoRegC except for MCE in which BBQ is superior to ENIR; however, the difference is not statistically significant for MCE. The results show that in terms of discrimination based on AUC, there is not a statistically significant difference between the performance of ENIR compared with BBQ and IsoRegC. However, ENIR performs statistically better than BBQ in terms of ACC. In terms of calibration measures, ENIR is statistically superior to both IsoRegC and BBQ in terms of RMSE. In terms of MCE, ENIR is statistically superior to IsoRegC.

% However, ENIR still performs statistically equivalent to BBQ in terms of MCE. 

Table \ref{tab:SVM} shows the results when we use SVM as the base classifier. As shown, the performance of ENIR is always superior to BBQ and IsoRegC except for MCE in which BBQ performs better than ENIR; however, the difference is not statistically significant for MCE. The results show that although ENIR is superior to IsoRegC and BBQ in terms of discrimination measures, AUC and ACC, the difference is not statistically significant. In terms of calibration measures, ENIR performs statistically superior to BBQ in terms of RMSE and it is statistically superior to IsoRegC in terms of MCE.

% However, ENIR and BBQ still perform statistically equivalently in terms of MCE. 

Table \ref{tab:NB} shows the results of comparison when we use NB as the base classifier. As shown, the performance of ENIR is always superior to BBQ and IsoRegC. In terms of discrimination, for AUC there is not a statistically significant difference between the performance of ENIR compared with BBQ and IsoRegC; however, in terms of ACC, ENIR is statistically superior to BBQ. In terms of calibration measures, ENIR is always statistically superior to IsoRegC. ENIR is also statistically superior to BBQ in terms of ECE and RMSE.

Overall, in terms of discrimination measured by AUC and ACC, the results show that the proposed calibration method either outperforms IsoRegC and BBQ, or has a performance that is not statistically significantly different. In terms of calibration measured by ECE, MCE, and RMSE, ENIR either outperforms other calibration methods, or it has a statistically equivalent performance to IsoRegC and BBQ.

% Table generated by Excel2LaTeX from sheet 'Real Data'
\begin{table}[htbp]
  \centering
  \small
    \begin{tabular}{rlll}
    \toprule
    \multicolumn{1}{c}{} & IsoReg & BBQ   & ENIR   \\
    \midrule
    \multicolumn{1}{l}{AUC}     & 1.963         & 2.225             & \textbf{1.813}     \\
    \multicolumn{1}{l}{ACC}     & 1.675         & 2.663$\ast$       & \textbf{1.663}     \\
    \multicolumn{1}{l}{RMSE}    & 1.925$\ast$   & 2.625$\ast$       & \textbf{1.450}     \\
    \multicolumn{1}{l}{ECE}     & 2.125         & 1.975             & \textbf{1.900}     \\
    \multicolumn{1}{l}{MCE}     & 2.475$\ast$   & \textbf{1.750}    & 1.775              \\
    % \multicolumn{1}{l}{LL}      & 2.400$\ast$   & 1.850             & \textbf{1.750}    \\
    \bottomrule
    \end{tabular}%
    \caption{Average rank of the calibration methods on the benchmark datasets using LR as the base classifier. Marker $\ast$/$\circledast$ indicates whether ENIR is statistically superior/inferior to the compared method (using the Friedman test followed by Holm's step-down procedure at a $0.05$ significance level).}
  \label{tab:LR}%
\end{table}% 

% Table generated by Excel2LaTeX from sheet 'Real Data'
\begin{table}[htbp]
  \centering
  \small
    \begin{tabular}{rlll}
    \toprule
    \multicolumn{1}{c}{} & IsoReg & BBQ   & ENIR   \\
    \midrule
    \multicolumn{1}{l}{AUC}     & 1.988         & 2.025             & \textbf{1.988}     \\
    \multicolumn{1}{l}{ACC}     & 2.000         & 2.150             & \textbf{1.850}     \\
    \multicolumn{1}{l}{RMSE}    & 1.850         & 2.475$\ast$       & \textbf{1.675}     \\
    \multicolumn{1}{l}{ECE}     & 2.075         & 2.025             & \textbf{1.900}    \\
    \multicolumn{1}{l}{MCE}     & 2.550$\ast$   & \textbf{1.625}    & 1.825              \\
    % \multicolumn{1}{l}{LL}      & 2.300$\ast$   & 1.975             & \textbf{1.725}    \\
    \bottomrule
    \end{tabular}%
    \caption{Average rank of the calibration methods on the benchmark datasets using SVM as the base classifier. Marker $\ast$/$\circledast$ indicates whether ENIR is statistically superior/inferior to the compared method (using the Friedman test followed by Holm's step-down procedure at a $0.05$ significance level).}
  \label{tab:SVM}%
\end{table}% 

% Table generated by Excel2LaTeX from sheet 'Real Data'
\begin{table}[htbp]
  \centering
  \small
    \begin{tabular}{rlll}
    \toprule
    \multicolumn{1}{c}{} & IsoReg & BBQ   & ENIR   \\
    \midrule
    \multicolumn{1}{l}{AUC}     & 2.150         & 1.925             & \textbf{1.925}  \\
    \multicolumn{1}{l}{ACC}     & 1.963         & 2.375$\ast$       & \textbf{1.663}  \\
    \multicolumn{1}{l}{RMSE}    & 2.200$\ast$   & 2.375$\ast$       & \textbf{1.425}  \\
    \multicolumn{1}{l}{ECE}     & 2.475$\ast$   & 2.075$\ast$       & \textbf{1.450}  \\
    \multicolumn{1}{l}{MCE}     & 2.563$\ast$   & 1.850             & \textbf{1.588}  \\
    % \multicolumn{1}{l}{LL}      & 2.088         & \textbf{1.925}    & 1.988           \\
    \bottomrule
    \end{tabular}%
    \caption{Average rank of the calibration methods on the benchmark datasets using NB as the base classifier. Marker $\ast$/$\circledast$ indicates whether ENIR is statistically superior/inferior to the compared method (using the Friedman test followed by Holm's step-down procedure at a $0.05$ significance level). }
  \label{tab:NB}%
\end{table}% 

In addition to comparing the performance of ENIR with IsoRegC and BBQ, we also show in Table \ref{tab:CI} the $95\%$ confidence interval for the mean of the random variable $X$, which is defined as the percentage of the gain (or loss) of ENIR with respect to the base classifier:
\begin{equation}
X = \frac{measure_{enir} - measure_{method}}{measure_{method}},
\end{equation}
where $measure$ is one of the evaluation measures AUC, ACC, ECE, MCE, or RMSE. Also, $method$ denotes one of the choices of the base classifiers, namely, LR, SVM, or NB. For instance, Table \ref{tab:CI} shows that by post-processing the output of SVM using ENIR, we are $95\%$ confident to gain 
anywhere from $17.6\%$ to $31\%$ average improvement in terms of RMSE. This could be a promising result, depending on the application, considering the $95\%$ CI for the AUC which shows that by using ENIR we are $95\%$ confident not to loose more than $1\%$ of the SVM discrimination power in terms of AUC (Note, however, that the CI includes zero, which indicates that there is not a statistically significant difference between the performance of SVM and ENIR in terms of AUC). 

Overall, the results in Table \ref{tab:CI} show that there is not a statistically meaningful difference between the performance of ENIR and the base classifiers in terms of AUC. The results support at a $95\%$ confidence level that ENIR improves the performance of LR or NB in terms of ACC. Furthermore, the results in Table \ref{tab:CI} show that by post-processing the output of LR, SVM, and NB using ENIR, we can make dramatic improvements in terms of calibration measured by RMSE, ECE, and MCE. For instance, the results indicate that at a $95\%$ confidence level, ENIR improved the average performance of NB in terms of ECE anywhere from $30.5\%$ to $55.2\%$, which could be practically significant in many decision-making and data mining applications.

% Table generated by Excel2LaTeX from sheet 'Real Data'
\begin{table}[htbp]
  \centering
  \small
    \begin{tabular}{rcccl||l}
    \toprule
    \multicolumn{1}{c}{} & LR & SVM   & NB   \\
    \midrule
    \multicolumn{1}{l}{AUC}  &[-0.008 , 0.003]     &  [-0.010 , 0.003]    & [-0.010 , 0.000] \\
    \multicolumn{1}{l}{ACC}  &[0.002 , 0.016]      &  [-0.001 , 0.010]    & [0.012 , 0.068] \\
    \multicolumn{1}{l}{RMSE} &[-0.124 , -0.016]    &  [-0.310 , -0.176]    & [-0.196 , -0.100] \\
    \multicolumn{1}{l}{ECE}  &[-0.389 , -0.153]    &  [-0.768 , -0.591]    & [-0.514 , -0.274] \\
    \multicolumn{1}{l}{MCE}  &[-0.313 , -0.064]    &  [-0.591 , -0.340]    & [-0.552 , -0.305] \\
    % \multicolumn{1}{l}{LL} &[ -0.106 ,	0.072]   &   [-0.432 , -0.230]    & [-0.522 , -0.030] \\
    \bottomrule
    \end{tabular}%
    \caption{ The $95\%$ confidence interval for the average percentage of improvement over the base classifiers(LR, SVM, NB) by using the ENIR method for post-processing. Positive entries for AUC and ACC mean ENIR is on average performing better discrimination than the base classifiers Negative entries for RMSE, ECE, and MCE mean that ENIR is on average performing better calibration than the base classifiers.}
  \label{tab:CI}%
\end{table}%

Finally, Table \ref{tab:summary} shows a summary of the time complexity of  different binary classifier calibration methods in learning for N training instances and the test time for only one instance.

\begin{table}[htbp]
  \centering
  \small
    \begin{tabular}{lll}
    \toprule
          & Training Time & Testing Time  \\
    \midrule
    Platt & $O(N T)$ & $O(1)$   \\
    Hist  & $O(N \log N)$ & $O(\log B)$  \\
    IsoRegC & $O(N \log N)$ & $O(\log B)$  \\
    ACP   & $O(N \log N)$ & $O(N)$   \\
    % SBB   & $O(N^2)$ & $O(B)$  \\
    ABB   & $O(N^2)$ & $O(N^2)$  \\
    BBQ   & $O(N \log N)$ & $O(M \log N)$  \\
    ENIR  & $O(N \log N)$ & $O(M \log B)$  \\
    \bottomrule
    \end{tabular}%
  \caption{Note that N and B are the size of training sets and the number of bins found by the method respectively. T is the number of iteration required for convergence in Platt method and M is defined as the total number of models used in the associated ensemble model.}
  \label{tab:summary}%
\end{table}%

\section{Conclusion}
\label{sec:conclusion}

In this paper, we presented a new non-parametric binary classifier calibration method called \textit{ensemble of near isotonic regression} (ENIR) that generalizes the isotonic regression based calibration method (IsoRegC) in two ways. First, ENIR makes a more realistic assumption compared to IsoRegC by assuming that the transformation from the uncalibrated output of a classifier to calibrated probability estimates is approximately (but not necessarily exactly) a monotonic function. Second, ENIR is an ensemble model that utilizes the BIC scoring function to perform selective model averaging over a set of near isotonic regression models that indeed includes IsoRegC as an extreme member. The method is computationally tractable, as it runs in $O(N \log N)$ for $N$ training instances. It can be used to calibrate many different types of binary classifiers, including logistic regression, support vector machines, na\"{i}ve Bayes , and others.  Our experiments show that by post processing the output of classifiers using ENIR, we can gain high calibration improvement in terms of RMSE, ECE, and MCE, without losing any statistically meaningful discrimination performance.
Moreover, our experimental evaluation on a broad range of real datasets showed that ENIR outperforms IsoRegC and BBQ (i.e. a state-of-the-art binary classifier calibration method \cite{pakdaman2015obtaining}). 

An important advantage of ENIR over BBQ is that it can be extended to a multi-class and multi-label calibration models similar to what has done for the standard IsoRegC \cite{zadrozny2002transforming}. This is an area of our current research. We also plan to investigate theoretical properties of ENIR. In particular, we are interested to investigate theoretical guarantees regarding the discrimination and calibration performance of ENIR, similar to what has been proved for the AUC guarantees of IsoRegC \cite{fawcett2007pav}.

% \begin{abstract}
% %\boldmath
% The abstract goes here.
% \end{abstract}


% \fi
\IEEEpeerreviewmaketitle

\bibliographystyle{plain}
\bibliography{refs}

\end{document}